\documentclass[twoside]{article}

\usepackage{hyperref}       
\hypersetup{colorlinks,linkcolor=blue,
            citecolor=blue,
            urlcolor=magenta,
            linktocpage,
            plainpages=false}
\usepackage{url}            
\usepackage{amsfonts}       
\usepackage{nicefrac}       
\usepackage{microtype} 
\usepackage{amsmath, amsthm, amssymb}
\usepackage{thm-restate}
\usepackage[ruled,algo2e]{algorithm2e}
\usepackage{algpseudocode}
\usepackage{bbm}
\usepackage{graphicx}
\usepackage{mathtools}
\usepackage{enumitem}
\usepackage{physics}
\usepackage{subcaption}
\usepackage{authblk} 
\usepackage{times}
\usepackage{tabulary, booktabs}
\usepackage{thmtools}
\usepackage{thm-restate}

\newtheorem{lem}{Lemma}

\newtheorem{rem}{Remark}

\newcommand{\thetabar}{{\overline{\theta}}}
\newcommand{\pibar}{{\bar{\pi}}}

\newcommand{\Bc}{\mathcal{B}}

\newcommand{\Sc}{\mathcal{S}}

\newcommand{\Ac}{\mathcal{A}}

\newcommand{\R}{\mathbb{R}}
\newcommand{\E}{\mathbb{E}}

\DeclareMathOperator*{\argmin}{arg\,min}

\renewenvironment{abstract}
  {{\centering\large\bfseries Abstract\par}\vspace{0.7ex}%
    \bgroup
       \leftskip 20pt\rightskip 20pt\small\noindent\ignorespaces}%
  {\par\egroup\vskip 0.25ex}

{\par\egroup\vskip 0.25ex}

\usepackage[accepted]{aistats2021}
\usepackage{natbib}

\begin{document}

%

%

\twocolumn[

\aistatstitle{On the Linear Convergence of Policy Gradient Methods for Finite MDPs}

\aistatsauthor{Jalaj Bhandari \And Daniel Russo}

\aistatsaddress{Columbia University \\ Simons Institute, UC Berkeley \And  
Columbia University} ]

\begin{abstract}
We revisit the finite time analysis of policy gradient methods in the one of the simplest settings: finite state and action MDPs with a policy class consisting of all stochastic policies and with exact gradient evaluations. There has been some recent work viewing this setting as an instance of smooth non-linear optimization problems and showing sub-linear convergence rates with small step-sizes. Here, we take a  different perspective based on connections with policy iteration and show that many variants of policy gradient methods succeed with large step-sizes and attain a linear rate of convergence. 
\end{abstract}

\section{Introduction}
Policy gradient methods, dating back to the works of \citep{williams1992simple,baxter1999direct,sutton2000policy,marbach2001simulation}, along with their modern variants \citep{kakade2002natural, silver2014deterministic}, have emerged as one of the most effective classes of algorithms for solving challenging reinforcement learning problems with impressive empirical success \citep{schulman2015trust,schulman2017proximal}. Despite this, little was known about their global convergence properties, as these methods search over a parameterized class of policies by performing (stochastic) gradient descent on a scalar loss function that is typically non-convex. 

This has changed recently with several recent papers analysing the global convergence properties of policy gradient methods. Our earlier work identifies properties  for general MDPs which guarantee that (despite non-convexity) the optimization landscape does not suffer from spurious local optima, thereby implying convergence of policy gradient methods to globally optimal solutions \citep{bhandari2019global}. Though that work does not consider specific algorithms, some convergence rates for follow easily from the framework (e.g. a sub-linear convergence rate for tabular MDPs using projected gradient descent with natural parameterization). The most comprehensive analysis of convergence rates appears in \cite{agarwal2019optimality}, showing results for different combinations of policy parametrization (natural and softmax policies), algorithms (projected and natural gradient descent) as well as entropy regularization\footnote{\cite{agarwal2019optimality} also go beyond tabular MDPs to give results for a \textit{compatible} function approximation setting}. \cite{shani2019adaptive} focus on analyzing trust region optimization methods \citep{schulman2015trust, schulman2017proximal} based on mirror descent \citep{beck2003mirror}, giving rates for both unregularized and regularized tabular MDPs. Essentially all of these papers view policy optimization as instances of general smooth nonlinear optimization problems. The analyses suggest small step-sizes to control for the error due to local linearization and show convergence to an $\epsilon$--optimal policy within either $O\left(\frac{1}{\epsilon}\right)$ or $O\left(\frac{1}{\epsilon^2}\right)$ iterations, depending on the precise algorithm used.

In this work, we revisit the finite time analysis of policy gradient methods in the simplest setting: finite state and action MDPs with a policy class consisting of all stochastic policies and with exact gradient evaluations. This setting was covered in the aforementioned works of \cite{bhandari2019global, agarwal2019optimality,shani2019adaptive}. Instead of viewing the problem through the lens of nonlinear optimization, we take a policy iteration perspective. We highlight that many forms of policy gradient can work with extremely large stepsizes and attain a \emph{linear} rate of convergence, meaning they require only $O(\log(1/\epsilon))$ iterations to reach an $\epsilon$--optimal policy. At the core of our ideas is a connection between policy gradients and policy iteration, which underlies the analysis in \cite{bhandari2019global}. 

For finite MDPs, we show that this leads to an extremely simple analysis  covering many different first-order methods applied to the policy gradient objective, including projected gradient descent, Frank-Wolfe, mirror descent, and natural gradient descent. In an idealized setting where step-sizes are set by line search, a one paragraph proof applies to all algorithms. For natural gradient algorithms, a slightly longer calculation studies a specific step-size sequence. In the final section of this paper, we also discuss a setting of approximate line search as well as natural gradient methods with entropy regularization.

\paragraph{Scope and purpose of this work:}
It is possible that readers might find our setting of tabular MDPs with access to exact gradients somewhat limited. It is worth noting that recent works of \citep{agarwal2019optimality,shani2019adaptive,cen2020fast,mei2020global} have all compared the convergence rates of different policy gradient methods in this setting. Our work clarifies that with exact gradient evaluations, much faster convergence rates can be achieved with larger step-sizes. The results on line search based step-size selection are especially idealized, but show that classical non-linear optimization techniques would automatically select larger step sizes and attain linear convergence rates. 

Small step-sizes may be critical for controlling approximation errors and stabilizing algorithms in practical settings. Studying such issues likely requires a model that focuses on approximation errors and incomplete policy classes. Our work instead offers a clear understanding of what to expect in a setting without these challenges.



\paragraph{On concurrent work:}

We remark on the concurrent works of \citep{cen2020fast, mei2020global} which also show linear convergence of exact policy gradient methods for entropy regularized tabular MDPs with softmax policies and exact gradients. The main motivation behind these works is to theoretically  characterize the  benefits of using entropy based regularizers to obtain faster convergence rates. While both analyze different variants (simple gradients vs natural gradients), using entropy regularization seems crucial to their results. Another key difference is that unlike \citep{cen2020fast,mei2020global}, our proof techniques rely on a direct connection between policy gradients and policy iteration, leading to concise proofs that are applicable to a broad range of algorithms along with transparent bounds with a clear dependence on all relevant constants. Instead of leveraging sophisticated algebra, our focus is on giving readers a clear understanding.
 

\section{Problem Formulation}
Consider a Markov decision process (MDP), which is a six-tuple $\left(\Sc, \Ac, g, P, \gamma, \rho \right)$, consisting of a state space $\Sc$, action space $\Ac$, cost function $g$, transition kernel $P$, discount factor $\gamma \in (0,1)$ and initial distribution $\rho$. We assume the state space $\Sc$ to be finite and index the states as $\Sc = \{s_1,\cdots, s_n\}$. For each state $s\in \Sc$, we assume that there is a finite set of $k$ arms to choose from and take the action space, $\Ac = \Delta^{k-1}$ to be the set of all probability distributions over those $k$ arms. That is, any action $a \in \Ac$ is a probability vector where each component $a_i$ denotes the probability of taking the $i$-th action. The transition kernel $P$ specifies the probability $P(s'|s,a)$ of transitioning to a state $s'$ upon choosing action $a$ in state $s$. The cost function $g(s,a)\in \mathbb{R}$ denotes the instantaneous expected cost incurred when selecting action $a$ in state $s$. Cost and transition functions can be naturally extended to functions on the probability simplex by defining:
\begin{align}
\label{eq: tabular costs and transitions_ch3}
g(s,a) &= \sum_{i=1}^{k} g(s,e_i) \, a_i, && P(s'| s,a) &= \sum_{i=1}^{k} P(s'| s, e_i) \, a_i.
\end{align}
where $e_i$ is the $i$-th standard basis vector, representing one of the $k$ possible arms. We assume that costs are non-negative, meaning $ g(s,e_i) \geq 0 $ for all $s \in \Sc$ and $i \in \{1, \ldots, k\}$. This holds without loss of generality, as one can always add the same large constant to the cost of each state and action without changing the decision problem.  

\paragraph{Cost-to-go functions and Bellman operators.}
A stationary policy $\pi: \Sc \to \Ac$ selects a distribution over the $k-1$ dimensional simplex, $\Delta^{k-1}$ for each state $s \in \Sc$. We use the notation $\pi(s,i)$ to denote the probability of selecting action $i$ in state $s$ under policy $\pi$. Let $\Pi$ denote the set of all stationary policies over the simplex, 
\[
\Pi  = \{ \pi \in \mathbb{R}^{n\times k}_{+}  :  \sum_{i=1}^{k} \pi(s,i)  =1  \,\,\, \forall \, s\in \Sc \}.
\]
For any policy $\pi \in \Pi$, $J_{\pi}: \Sc \to \R$ is defined as,
\[
J_{\pi}(s) = \E_{\pi}\left[ \sum_{t=0}^{\infty} \gamma^t g(s_t, \pi(s_t)) \,\, \Big\vert \,\, s_0 = s \right].
\]
As the per-step costs are uniformly bounded, so are the cost-to-go functions. Define the Bellman operator $T_{\pi}: \mathbb{R}^n \to \mathbb{R}^n$ under policy $\pi$ and the Bellman optimality operator $T: \mathbb{R}^n \to \mathbb{R}^n$ as,
\begin{align*}
\left(T_{\pi} J\right)(s) &:= g(s, \pi(s)) + \gamma \sum_{s' \in \Sc} P(s' | s, \pi(s)) J(s') \\
\left(T J\right)(s) &:= \min_{a \in \Ac} \left[ g(s,a) + \gamma \sum_{s' \in \Sc} P(s'|s,a) J(s') \right].
\end{align*} 
Note that the Bellman optimality operator can be equivalently defined as $(TJ)(s) = \min_{\pi \in \Pi} (T_{\pi}J)(s)$. The cost-to-go function under policy $\pi$ is the unique solution to the Bellman equation, $J_\pi = T_{\pi} J_\pi$. Similarly, the optimal cost-to-go function, $J^*$ which satisfies $J^*(s) = \min_{\pi} J_{\pi}(s)$ for all $s\in \Sc$, is the unique fixed point of $T$ and that there is at least one optimal policy, $\pi^* \in \Pi$ that attains this minimum for every $s\in \Sc$. From the above definitions, it is simple to check that: $J_{\pi}=T_{\pi}J_{\pi} \succeq T J_{\pi}$ for any $\pi\in\Pi$. We will use this inequality repeatedly throughout our analysis.

Our analysis uses a few basic properties of Bellman operators, see \cite{bertsekas1995dynamic} or \cite{puterman2014markov} for proofs. Under the assumption that per-period costs are bounded, $T$ and $T_{\pi}$ are monotone, meaning the element-wise inequality $J \preceq J'$ implies $TJ\preceq TJ'$ and $T_{\pi}J \preceq T_{\pi}J'$. They are also contraction operators with respect to the maximum norm. That is,  $\|  TJ - TJ' \|_{\infty} \leq \gamma \|J - J' \|_{\infty}$ and  $\| T_{\pi} J - T_{\pi}J' \|_{\infty} \leq \gamma \|J - J' \|_{\infty}$ hold for for any $J,J' \in \mathbb{R}^n$. The state-action cost-to-go function under  policy $\pi\in \Pi$,
\[
Q_{\pi}(s,a)= g(s,a)+ \gamma \sum_{s' \in \Sc} P(s'\mid s,a) J_{\pi}(s'),
\]
measures the cumulative expected cost of taking action $a$ in state $s$ and applying $\pi$ thereafter. For any polices $\pi,\pi'\in \Pi$, we have the following relations:
\begin{align*}
Q_{\pi}(s,\pi(s)) &= J_{\pi}(s), \\
Q_{\pi}(s,\pi'(s)) &= (T_{\pi'} J_{\pi})(s), \\
\min_{a\in \Ac} Q_{\pi}(s,a) &= (TJ_{\pi})(s).
\end{align*}
Note that for any policy $\pi \in \Pi, s \in \Sc$ and $a \in \Delta^{k-1}$, linearity of the cost and transitions functions in \eqref{eq: tabular costs and transitions_ch3} implies that the Q-function is linear in $a$.
\[
Q_{\pi}(s,a) = \sum_{i=1}^k Q_{\pi}(s,e_i) a_i = \langle Q_{\pi}(s,\cdot), a \rangle
\]

\paragraph{Loss function and initial distribution.} Policy gradient methods seek to minimize the scalar loss function
\[
\ell(\pi)=  (1-\gamma)\sum_{s \in \Sc} J_{\pi}(s) \, \rho(s),
\]
in which the states are weighted by their initial probabilities under $\rho$ and we have normalized costs by $(1-\gamma)$ for convenience. We assume throughout that $\rho$ is supported on $\Sc$, meaning that $\rho(s)>0$ for all $s\in \Sc$ which implies that $\pi \in \argmin_{\bar{\pi}} \ell(\bar{\pi})$ if and only if $\pi\in\argmin_{\bar{\pi} } J_{\bar{\pi}}(s) \,\,\, \forall \, s \in \Sc$. Assuming an exploratory initial distribution is critical as it is well known that, in the absence of strong assumptions on the transition kernel, policy gradient methods can fail catastrophically if applied without some form of intelligent exploration. See \citep{thrun1992cient, kakade2002approximately} for a simple example and the discussions in \citep{agarwal2019optimality, bhandari2019global}.

\paragraph{State distributions.}  We define the discounted state occupancy measure under any policy $\pi$ and initial state distribution $\rho$ as:
\[
\eta_{\pi} = (1-\gamma)\sum_{t=0}^{\infty} \gamma^{t} \rho P_{\pi}^t = (1-\gamma) \rho (I - \gamma P_\pi)^{-1},
\]
where $\eta_{\pi}$ and $\rho$ are both row vectors, $P_{\pi} \in \mathbb{R}^{n\times n}$ denotes the Markov transition matrix  under $\pi$, i.e. $P_{\pi} = (P(s'| s, \pi(s)))_{s,s'\in \Sc}$ and $P_{\pi}^t$ denotes its $t$-step counterpart. Thus, $\eta_{\pi}$ is essentially the discounted fraction of time the system spends in a given state. Note that we have $\eta_{\pi}(s) \geq (1-\gamma) \rho(s) > 0$ as we assumed $\rho(s) > 0$ for all $s \in \Sc$.

\section{Linear convergence of policy iteration}
\label{sec: background}
We briefly revisit the classic policy iteration algorithm as our analysis of policy gradient methods is intricately tied to it. Starting from an initial policy $\pi$, policy iteration first evaluates the corresponding cost-to-go function $Q_{\pi}$, and then updates to a new policy $\pi^+$ such that
\[
\pi^+(s) \in \argmin_{a\in \Ac} \, Q_{\pi}(s,a) \quad \forall \, s\in \Sc.  
\]
In terms of the Bellman operators, this can be equivalently expressed as, $T_{\pi^+} J_{\pi} = TJ_{\pi}$. A simple analysis of policy iteration follows by using the monotonicity and contraction properties of the Bellman operators. Observe that
\begin{align}
\label{eq: bellman update reduces cost_ch3}  
J_{\pi} = T_{\pi}J_{\pi} \succeq T J_{\pi} = T_{\pi^+} J_{\pi}.
\end{align} 
Inductively applying $T_{\pi^+}$ to each side and using the monotonicity property yields a policy improvement property,
\begin{align}
\label{eq:policy_improvement_inequalities_ch3}
J_{\pi} \succeq T_{\pi^+} J_{\pi} \succeq T^{2}_{\pi^+} J_{\pi} \succeq \cdots \succeq J_{\pi^+}.
\end{align}
Here we use the definition that $J_{\pi^+} = \lim_{k \to \infty} \, T_{\pi^+}^k J$ for any $J \in \mathbb{R}^n$. Since $J_{\pi} \succeq TJ_{\pi} \succeq J_{\pi^+} \succeq J^*$ we have, 
\begin{align}
\label{eq:geomteric_contraction_PI_ch3}
\|  J_{\pi^+} - J^* \|_{\infty} \leq  \| TJ_{\pi} - J^* \|_{\infty} &= \| TJ_{\pi}- TJ^* \|_{\infty} \nonumber \\
&\leq \gamma \| J_{\pi} - J^*\|_{\infty},
\end{align}
using the contraction property. From this, we conclude that policy iteration converges to the optimal policy at a linear rate. Let $\{\pi^t\}_{t \geq 0 }$ be the set of policies produced by policy iteration. Then iterating over \eqref{eq:geomteric_contraction_PI_ch3} shows
\[ 
\| J_{\pi^t} -J^*\|_{\infty} \leq \gamma \| J_{\pi^{t-1}} - J^*\|_{\infty} \leq \cdots \leq \gamma^t \| J_{\pi^0} -J^*\|_{\infty}.
\] 
In fact, policy iteration can sometime also converge quadratically in the limit \citep{puterman2014markov}.

\section{A sharp connection between policy gradient and policy iteration}
Recently, \cite{bhandari2019global} analyze the optimization landscape of the policy gradient objective $\ell(\cdot)$ for general MDPs and policy classes. A starting point of that analysis is rewriting the policy gradient theorem in a form that emphasizes the illuminating connections between policy gradient and policy iteration.  We specialize that presentation to the tabular setting and argue that several first-order methods applied to the policy gradient loss $\ell(\cdot)$ will essentially perform a \textit{soft policy iteration} update and hence converge at a geometric rate, similar to policy iteration. 

For any policy $\pi \in \Pi$, consider the weighed policy iteration or "Bellman" objective, defined as 
\[
\Bc( \pibar | \eta_{\pi}, J_{\pi})= \sum_{s \in \Sc} \eta_{\pi}(s) Q_{\pi}(s, e_i) \pibar(s,i)  = \langle Q_{\pi} \, , \bar{\pi} \rangle_{\eta_{\pi} \times 1}
\]
where $e_i$ denotes the $i$-th standard basis vector, denoting one of the $k$ arms, $\langle v,u\rangle_{W} = \sum_{s=1}^{n}\sum_{i=1}^{k} v(s,i) u(s,i) W(s,i)$ denotes the $W$-weighted inner product and $\eta_\pi \times 1$ denotes a weighting that places weight $\eta_{\pi}(s) \cdot 1$ on any state-action pair $(s,i)$. Recall that since $\rho(s)>0$ by assumption, $\eta_{\pi}(s) > 0$ for all $s \in \Sc$ and hence the policy iteration update can be equivalently written as optimizing the Bellman objective,
\[
\pi^+ = \argmin_{\pibar \in \Pi} \Bc(\pibar | \eta_{\pi}, J_{\pi}).
\]
It is worth emphasizing that the Bellman cost function is a \textit{single period} objective, considering the cost-to-go of following $\pibar$ for a single period and following $\pi$ thereafter. A policy gradient theorem connects gradients of the infinite horizon cost function $\ell(\cdot)$ to gradients of the \textit{single period} Bellman objective underlying policy iteration. In particular, we have the following lemma from \cite{bhandari2019global}, which is essentially a restatement of the classical version by \citep{sutton2000policy, sutton2018reinforcement}.
\begin{lem}[Policy gradient theorem for tabular MDPs] 
\label{lem:pgtheorem}
Assuming per-period costs are uniformly bounded, $\ell(\pi)$ is continuously differentiable and 
\[
\frac{\partial \ell(\pi)}{\partial \pi(s,i)} = \frac{\partial \Bc(\pibar|\eta_{\pi},J_{\pi})}{\partial \pibar(s,i)} \bigg\vert_{\pibar = \pi} = \eta_{\pi}(s) Q_{\pi}(s, e_i) 
\] 
\end{lem}
Equivalently, we can write a first order Taylor expansion of $\ell(\cdot)$ as
\begin{align*}
\ell(\pibar) &= \ell(\pi) +  \langle \nabla \ell(\pi)\, ,\,  \pibar -\pi \rangle + O(\| \pibar - \pi \|^2 )\\
&= \ell(\pi) + \langle Q_{\pi} \, , \, \pibar-\pi  \rangle_{\eta_\pi \times 1 } + O(\| \pibar - \pi \|^2 ).
\end{align*}
Presentation of the policy gradient theorem in terms of the Bellman objective clarifies an important connection -- we can interpret $\nabla \ell(\pi)$ as  gradient of the weighted policy iteration objective.  What is special about the tabular setting, relative to the general problems considered by \cite{bhandari2019global}, is that the weighted policy iteration objective is \emph{linear}. In the following section, we use this connection to show that various first-order methods applied to $\ell(\cdot)$ can optimize the Bellman objective $\Bc(\cdot | \eta_{\pi}, J_{\pi})$ to optimality in a single update with large (and possibly infinite) step-sizes; equivalent to a policy iteration update. For finitely large step-sizes, a simple argument establishes equivalence between a policy gradient step and a soft policy iteration update, again implying geometric convergence.

Note that for tabular MDPs, a policy iteration step is simple as it reduces to solving a linear optimization problem over the probability simplex, and the optimal solution is to select the best action for each state.

\section{Policy gradient methods for finite MDPs}
\label{sec:algorithms}
We write all algorithms in terms of their evolution in the space of policies $\Pi$. Several of them could instead be viewed as operating in the space of parameters for some parameterized policy class. We discuss this in Remark \ref{rem:parameterization}, but keep our formulation and results focused on the space of policies $\Pi$. Note that $\Pi = \Delta^{k-1} \times \cdots \times \Delta^{k-1}$ is the $n$-fold product of the probability simplex. This form of the policy class will cause policy gradient updates to decouple across states. 

\begin{description}[leftmargin=0pt]
	\item[Frank-Wolfe.] Starting with some policy $\pi\in \Pi$, an iteration of the Frank-Wolfe algorithm computes
	\begin{align}
	\label{eq: tabular-franke-wolfe-optimization}
	\pi^+ = \argmin_{\pibar \in \Pi}  \, \langle \nabla\ell(\pi), \pibar \rangle =  \argmin_{\pibar \in \Pi} \, \langle Q_{\pi}, \pibar \rangle_{\eta_{\pi} \times 1} 
	\end{align}
	and then updates the policy to $\pi' =(1-\alpha) \pi + \alpha \pi^+$ for $\alpha \in [0,1]$. We use the notation $\pi^+$ in \eqref{eq: tabular-franke-wolfe-optimization} as it is exactly the policy iteration update to $\pi$ so \emph{Frank-Wolfe mimics a soft-policy iteration step}, akin to the \textit{conservative policy iteration} update\footnote{A generalized version of Frank-Wolfe was studied in \citep{scherrer2014local} under the name of ``Boosted Policy Search'' to show global optimality guarantees for any locally optimal policy. 
	} 
    in \cite{kakade2002approximately}. Note, the minimization problem in \eqref{eq: tabular-franke-wolfe-optimization} decouples across states to optimize a linear objective over the probability simplex, so 
	\[
	\pi^{+}(s) \in \argmin_{d \in \Delta^{k-1}} \, d^{\top} Q_{\pi}(s, \cdot)
	\]
	is a point-mass that places all weight on $\argmin_{i} Q_{\pi}(s, e_i)$. 
	\item[Projected Gradient Descent.] Starting with some policy $\pi\in \Pi$, an iteration of the projected gradient descent algorithm with constant stepsize $\alpha$ updates to the solution of the following regularized problem
	\begin{align*}
	   \pi' &= \argmin_{\pibar \in \Pi}  \langle \nabla\ell(\pi), \pibar \rangle + \frac{1}{2\alpha} \|  \pibar-\pi \|_2^2 \\
	   &=  \argmin_{\pibar \in \Pi} \langle Q_{\pi}, \pibar \rangle_{\eta_{\pi} \times 1}  + \frac{1}{2\alpha} \|  \pibar-\pi \|_2^2.
	\end{align*}
	As $\alpha \to \infty$ (the regularization term tends to zero), $\pi'$ converges to the solution of \eqref{eq: tabular-franke-wolfe-optimization}, which is exactly the policy iteration update as noted above. For intermediate values of $\alpha$, the projected gradient update decouples across states and takes the form: 
	\[
	\pi'(s)  = {\rm Proj}_{2, \Delta^{k-1}}(\pi(s) - \alpha Q_{\pi}(s, \cdot))\]
	which is a gradient step followed by projection onto the probability simplex. Note that from an implementation perspective, projections onto the probability simplex involves a computationally efficient ($\mathcal{O}(k \log k)$) soft-thresholding operation \citep{duchi2008efficient}. 
	\item[Mirror-descent.] The mirror descent method adapts to the geometry of the probability simplex by using a non-euclidean regularizer. We focus on using the Kullback Leibler (KL) divergence, a natural choice for the regularizer, under which an iteration of mirror descent updates policy $\pi$ to $\pi'$ as: 
	\begin{align*}
	\pi' &= \argmin_{\pibar \in \Pi}  \langle \nabla\ell(\pi), \pibar \rangle + \frac{1}{\alpha} \sum_{s=1}^{n} D_{\rm KL}( \pibar_s \, || \, \pi_s) \\
	&= \argmin_{\pibar \in \Pi} \langle Q_{\pi}, \pibar \rangle_{\eta_{\pi} \times 1}  + \frac{1}{\alpha}\sum_{s=1}^{n} D_{\rm KL}( \pibar_s \, || \, \pi_s),
	\end{align*}
	where $D_{\rm KL}(p||q)=\sum_{i=1}^{k} p_i\log(p_i /q_i)$ denotes the KL divergence. It is well know that the solution to this optimization problem is the exponentiated gradient update \cite[Section 6.3]{bubeck2015convex},
	\begin{equation}
	\label{eq:mirror_descent_update}
	\pi'(s,i) = \frac{\pi(s,i) \cdot \exp\{ -\alpha \eta_{\pi}(s) Q_{\pi}(s,e_i) \} }{ \sum_{j=1}^{k} \pi(s,j) \cdot \exp\{ -\alpha \eta_{\pi}(s) Q_{\pi}(s,e_j) \}}. 
	\end{equation}
	Again, we can see that $\pi'$ converges to a policy iteration update as $\alpha \to \infty$.
	
	\item[Natural policy gradient and TRPO.] We consider the natural policy gradient (NPG) algorithm of \cite{kakade2002natural} which is closely related to the widely used TRPO algorithm of \cite{schulman2015trust}. We focus on NPG applied to the \textit{softmax parameterization} for which it is actually an instance of mirror descent with a specific regularizer. In particular, beginning with some policy $\pi \in \Pi$, an iteration of NPG updates to $\pi'$:
    \begin{align}
    \pi'&=\argmin_{\pibar \in \Pi} \,\, \langle \nabla\ell(\pi), \pibar \rangle + \frac{1}{\alpha} \sum_{s=1}^{n} \eta_{\pi}(s) D_{\rm KL}( \pibar_s \, || \, \pi_s ) \nonumber \\
    &= \argmin_{\pibar \in \Pi} \,\, \langle Q_{\pi}, \pibar \rangle_{\eta_{\pi} \times 1}  + \frac{1}{\alpha}\sum_{s=1}^{n} \eta_{\pi}(s) D_{\rm KL}( \pibar_s \, || \, \pi_s  ) \label{eq:NPGupdateform}
    \end{align}
	using a regularizer that penalizes changes to the action distribution at states in proportion to their occupancy measure $\eta_{\pi}$. As discussed above, it is well known that this KL divergence regularized problem is solved by an exponentiated weights update for each state $s\in \{1,\ldots, n \}$,
	\begin{align}
	\label{eq:NPG_update}
	 \pi'(s,i) = \left(\frac{\pi(s,i) \cdot \exp\{ -\alpha  Q_{\pi}(s,e_i) \} }{ \sum_{j=1}^{k} \pi(s,j) \cdot \exp\{ -\alpha Q_{\pi}(s,e_j)  \} }\right).
	\end{align}
	Note that as compared to \eqref{eq:mirror_descent_update}, this update rule is independent of the state occupancy measure $\eta_{\pi}$. A potential source of confusion is that natural policy gradient is usually described as steepest descent in a variable metric defined by the Fisher information matrix induced by the current policy\footnote{This is equivalent to
	mirror descent under some conditions \citet{raskutti2015information}.},
    \begin{align*}
        \pi' &= \pi + \alpha F(\pi)^{\dagger} \nabla \ell(\pi) \\
        F(\pi) &= \sum_{s,i} \eta_{\pi}(s) \pi(s,i) \left[ \nabla \log \pi(s,i) \left( \nabla \log \pi(s,i) \right)^\top \right]
    \end{align*}
	where $M^{\dagger}$ denotes the pseudoinverse of matrix $M$. Readers can check that the exponentiated update in \eqref{eq:NPG_update} matches the explicit formula for the NPG update with softmax policies as given in \cite{kakade2002natural} and \cite{agarwal2019optimality}.  
\end{description}

Step-size selection is an important issue for most first order methods. Each of the algorithms above can be applied with a sequence of stepsizes $\{\alpha_{t}\}_{t \geq 0}$ to produce a sequence of policies $\{\pi^t \}_{t \geq 0}$.  We define one stepsize selection rule below. 

\paragraph{Exact line search.}
At iteration $t$, the update rules for each of the algorithms described above actually specify a new policy $\pi^{t+1}_{\alpha}$ for a range of stepsizes, $\alpha \geq 0$. We consider an idealized stepsize rule using \emph{exact line search}, which directly optimizes over this choice of stepsize at each iteration, selecting  $\pi^{t+1} = \pi^{t+1}_{\alpha^*}$ where $\alpha^* = \argmin_{\alpha} \ell(\pi^{t+1}_{\alpha})$ whenever this minimizer exists. More generally, we define
\begin{align}
\label{eq: exact linesearch}
\pi^{t+1} = \argmin_{\pi \in \Pi^{t+1}}  \ell(\pi).
\end{align}
where $\Pi^{t+1} = {\rm Closure}(\{ \pi_{\alpha}^{t+1}\})$ denotes the closed curve of policies traced out by varying $\alpha$. For Frank-Wolfe, $\Pi^{t+1} = \{ \alpha \pi^t + (1-\alpha) \pi^t_+ : \alpha \in [0,1]\} $ is the line segment connecting the current policy $\pi^t$ and its policy iteration update $\pi^t_+$. Under NPG, $\{\pi^{t+1}_{\alpha} \}$ is a curve where $\pi^{t+1}_{0}=\pi^t$ and $\pi^{t+1}_{\alpha} \to \pi^t_+$ as $\alpha \to \infty$. Since $\pi^t_+$ is not attainable under any fixed $\alpha$, this curve is not closed. By taking the closure, and defining line search via \eqref{eq: exact linesearch}, certain formulas become cleaner. Of course, it is also possible to nearly solve \eqref{eq: exact linesearch} without taking the closure and obtain essentially the same results. We elaborate on this in the discussion that follows our main result in Theorem \ref{thm: tabular rates}.   

\begin{rem}[Policy parameterization and infima vs minima]\label{rem:parameterization}
We chose to work with the class of all stochastic policies $\Pi$ (often termed as \textit{natural parameterization}) as opposed to some parameterized policy classes, which are more commonly used in practice. For example, a policy gradient algorithm might search over the parameter $\theta \in \mathbb{R}^{n\times k}$ of a softmax policy $\pi_{\theta} \in \Pi$, defined by $\pi_{\theta}(s,i) \propto e^{\theta_{s,i}}$. For example, consider the  TRPO algorithm proposed by  \citep{schulman2015trust} which uses a locally linearization of $\ell(\pi)$, forms the regularized minimization problem in \eqref{eq:NPGupdateform}, and then updates the parameter of a softmax policy $\pi_{\theta}$ by solving
\[
\argmin_{\thetabar} \,\, \langle Q_{\pi_{\theta}}, \pi_{\thetabar} \rangle_{\eta_{\pi_{\theta}} \times 1}  + \frac{1}{\alpha}\sum_{s=1}^{n} \eta_{\pi_{\theta}}(s) D_{\rm KL}( \pi_{\thetabar}(s) \, || \, \pi_{\theta}(s)).
\]
We could define similar versions of projected gradient descent or Frank-Wolfe, which also linearize $\ell(\pi)$, but then optimize the resulting local approximation only over parameterized policies. Since the class of softmax policies can approximate any stochastic policy to arbitrary precision, this is nearly the same as optimizing over the policy class $\Pi$. Studying $\Pi$ directly makes mathematical analysis easier, because it is closed. For example, it contains an optimal policy, whereas any softmax policy $\pi_{\theta}$ can only come infinitesimally close to an optimal policy. In practice, optimization problems are never solved beyond machine precision, so we don't view the distinction between infimum and minimum to be relevant to the paper's main insights.  We caution the reader that our results do not apply to more naive gradient methods that directly linearize $\ell(\pi_{\theta})$ with respect to $\theta$.  In that case,  a gradient update to $\theta$ may not approximate a policy iteration update, no matter how large the stepsize is chosen to be. In fact, such methods may perform badly due to issues of poor conditioning \citep{kakade2002natural}. 
 


\end{rem} 

\section{Main result: geometric convergence}
So far, we have described different variants of  policy gradient methods for tabular MDPs. For large step-sizes, all these algorithms essentially make a policy iteration update. Hence, intuitively, it is reasonable to expect that their convergence behavior closely resembles that of policy iteration rather than that of gradient descent for smooth objectives. We quantify this precisely in Theorem \ref{thm: tabular rates} below.

Our first result confirms that all of the algorithms we presented in the previous section converge geometrically when step-sizes are set by exact line search on $\ell(\cdot)$. Again, the idea is that \textit{a policy gradient step is a policy iteration update} for an appropriate choice of stepsize. Our proof effectively uses that exact line search updates make at least as much progress in reducing $\ell(\cdot)$ as a policy iteration update.
The mismatch between the policy gradient loss $\ell(\cdot)$, which governs the stepsize choice, and the maximum norm, which governs policy iteration convergence, is the source of the term $\min_{s\in \Sc} \rho(s)$ in the bound. We further elaborate on this issue in the discussion that follows Theorem \ref{thm: tabular rates}.

Our second and third results show that dependence on the initial distribution in the bounds can be avoided by forcing the algorithm to use large stepsizes. A simple result in part (b) applies to the Frank-Wolfe algorithm with a constant stepsize, which gives performance improvement in max norm. This bound follows by essentially making a minor modification to the linear convergence result of policy iteration as reviewed in Section \ref{sec: background}. Recall that we already showed a Frank-Wolfe update to be exactly equivalent to a soft policy iteration update, 
\[
\pi^{t+1}(s) = (1-\alpha) \pi^t(s) + \alpha \pi^t_+(s).
\]
Given this close connection, a simple argument shows that an $\alpha$-step Frank-Wolfe update offers at least a fraction of the performance improvement offered by a policy iteration update,
\[
J_{\pi^{t+1}} \preceq (1-\alpha) J_{\pi^t} + \alpha T J_{\pi^t}
\]
which implies the result. A comparison between parts (a) and (b) suggest that for $\alpha \geq 1/|\Sc|$, Frank-Wolfe with exact line search might converge slowly as compared to the constant step-size version in the worst case.

For softmax policies and exact gradient evaluations, we show in part (c) that NPG with an \textit{adaptive step-size sequence} converges to an $\epsilon$ optimal policy in $O(\log(1/\epsilon))$ iterations. The error term, $\epsilon$, is inversely related to the step-size and reflects the fact that NPG updates with finite step-sizes only approximately resemble the policy iteration updates\footnote{More precisely, our proof shows that in this case, the NPG update is equivalent to a soft policy iteration update upto some additive error.}. As we take the step-size to infinity, we recover the same result as one would expect for policy iteration. Compared to the first result in part (a) which applies with exact line search, the result in part (c) is useful in the sense that it gives a precise quantification of how large the step-sizes need to be for linear convergence to hold.

\begin{restatable}[Geometric convergence]{thm}{geometricthm}
	\label{thm: tabular rates}
	Suppose one of the first-order algorithms in Section \ref{sec:algorithms} is applied to minimize $\ell(\pi)$ over $\pi \in \Pi$ with step-size sequence $\{\alpha_t\}_{t\geq0}$. Let $\pi^0$ denote the initial policy and $\{\pi^{t}\}_{t\geq0}$ denote the sequence of iterates. The following bounds apply.
	\begin{enumerate}[label=(\alph*)]
		\item  \textbf{Exact line search.} If either Frank-Wolfe, projected gradient descent, mirror descent, or NPG is applied with step-sizes chosen by exact line search as in \eqref{eq: exact linesearch}, then
		\[ 
		\|  J_{\pi^t} - J^* \|_{\infty}  \leq  \Big(1- \min_{s \in \Sc} \rho(s) ( 1-\gamma)\Big)^t  \frac{\|  J_{\pi^0} - J^* \|_{\infty} }{\min_{s \in \Sc} \rho(s)}. 
		\] 
		\item \textbf{Constant step-size Frank-Wolfe.} Under Frank-Wolfe with constant step-size $\alpha\in (0,1]$, 	\[ 
		  \|  J_{\pi^t} - J^* \|_{\infty}  \leq (1-\alpha(1-\gamma))^t \|  J_{\pi^0} - J^* \|_{\infty}. 
		\]
		\item \textbf{Natural policy gradient with softmax policies and adaptive step-size.} Fix any $\epsilon>0$. Let $i^*_t = \argmin_{i} Q_{\pi^t}(s,i)$. Suppose that NPG is performed with an adaptive step-size sequence,
		\[\alpha_t(s) \geq  \frac{2}{(1-\gamma) \epsilon} \log(\frac{2}{\pi^t(s,i^*_t)}). \] 
		Then, 
		\[
		\norm{J_{\pi^t}-J^*}_{\infty} \leq \left(\frac{1+\gamma}{2} \right)^t \norm{J_{\pi^0}-J^*}_{\infty} + \epsilon.
		\]
	\end{enumerate}
    \begin{rem}
        For the result in part (c), note that for the softmax parameterization, $\pi_{\theta}(s,i) > 0$ for any $\theta \in \mathbb{R}^{n \times k}$. So, $\pi^t(s,i^*_t) > 0$ for all $t$. A similar result can also be obtained without the need of adaptive step-sizes by considering entropy regularized MDPs. This is discussed below.
        \end{rem}
\end{restatable}

\paragraph{Discussion of results:} The following discussion is based primarily on feedback of the reviewers. We thank them for their valuable inputs. 
\begin{enumerate}
    \item \underline{Dependence on $\rho_{\min}$ for exact line search result}: \\
    Readers will note that proof of our result in part (a) of Theorem \ref{thm: tabular rates} also shows that,
    \begin{equation*}
    \ell(\pi^{t+1}) - \ell(\pi^*) \leq \left(1 - \rho_{\min} (1-\gamma)\right)^t \left[ \ell(\pi^0) - \ell(\pi^*) \right].
    \end{equation*}
    where $\rho_{\min} = \min_{s \in \Sc} \rho(s)$. A natural question to ask is whether the presence of the factor of $\rho_{\min}$ in the geometric rate is merely an artifact of our analysis technique and if in practice, line search always ends up picking the policy iteration update corresponding to $\alpha=1$. In Figure \ref{fig:fw_grid} above, we plot the line search objective, \[\{\ell(\pi_{\alpha}) : \pi_{\alpha} = \alpha \pi + (1-\alpha) \pi_{+}, \, \alpha \in [0,1]\},
    \]
    for a Frank-Wolfe update for a randomly generated\footnote{We generated many random MDPs to compare updates of policy iteration with those of Frank-Wolfe using grid search and found many cases where these differ. Details of only one such example is given to illustrate our point.} MPD with two states and three actions. For a given choice of $(P, g, \rho)$ and policy $\pi$ (see Appendix \ref{appendix_1} for details), we observe that $\ell(\pi_{\alpha})$ is non-monotonic in $\alpha$ and therefore exact line-search does often select smaller step-sizes as compared to the greedy update $(\alpha=1)$. Although we do not show a lower bound, this example suggests that a factor of $\rho_{\min}$ in the bound here might be unavoidable.
    \begin{figure}
	\centering
	\includegraphics[width=0.45\textwidth]{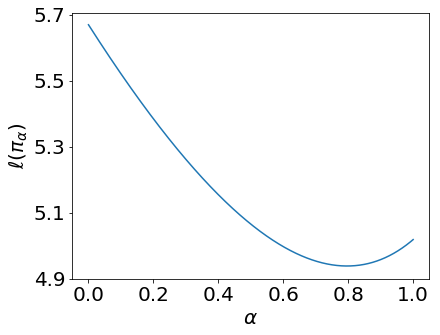}
	\caption{Line search objective for a Frank-Wolfe update for a two state three action MDP is non-monotonic with a minimum at $\alpha = 0.83$. Therefore, exact line search picks a smaller step-size than the greedy update, i.e. $\pi_{\alpha^*}\neq\pi_{+}$. }
	\label{fig:fw_grid}
\end{figure}
    \item \underline{On inexact line search}: Though our result in part (a) of Theorem \ref{thm: tabular rates} focuses on an idealized setting with exact line search, we do note that a similar result can also be obtained if we can ensure, say using inexact line search, that the improvement in total cost $\ell(\cdot)$ at every update is at least a fraction of the improvement offered by exact line search. For example, if we select a step-size sequence $\{\alpha_t\}_{t \geq 0}$ which offers half the possible improvement at every update, meaning $\ell(\pi^t) - \ell(\pi^{t+1}_{\alpha_t}) \geq (1/2)(\ell(\pi^t) - \inf_{\alpha'} \ell(\pi^{t+1}_{\alpha'}))$, then our result in part (a) follows with an extra factor of $\frac{1}{2}$ in the bound. One essentially needs to modify the first step in the proof (Equation \eqref{eq:exact_vs_inexact_linesearch}) and the rest is same. 
    
    A linear convergence result can also be obtained if the sequence of policies, $\{\pi^t\}_{t \geq 0}$, obtained via inexact line search offer approximately the same improvement as a policy iteration update, i.e. $\ell(\pi^{t+1}) \leq \ell(\pi^t_+) + \delta$ holds uniformly for some $\delta>0$. In this case, a bound similar to that in part (a) will hold with an additional scaled bias term of $\delta/(1-\gamma)$.
    
    \item \underline{NPG with softmax policies for regularized MDPs:} \\
    Recall that the result in part (c) uses an adaptive step-size sequence $\alpha_t(s)$ that depends on $\pi^t(s,i^*_t)$, probability under the randomized policy at iteration $t$ assigned to the action $i^*_t$ prescribed by policy iteration. This dependence is a bit undesirable and can be removed by considering \textit{entropy regularized} MDPs. Entropy regularization prevents policies from picking near deterministic actions and essentially lower bounds $\pi^t(s,i^*_t)$. Rather than presenting a lengthy re-derivation of the result in part (c), we sketch a simple argument essentially based on some past work on the theory of regularized MDPs \citep{neu2017unified,geist2019theory}, to show linear convergence with a particular choice of step-size. Although this result in Equation \eqref{eq:regmdpresult} is almost identical to the one in \cite{cen2020fast}, our ideas, based on connections to policy iteration, considerably simplify the proof. 
    
    A common way to enforce regularization is by adding a small penalty to the cost function, 
    \[
    g^{\lambda}(s,a) = \sum_{i=1}^k \left(g(s,e_i)a_i + \lambda \log(a_i)\right) 
    \]
    for some parameter $\lambda > 0$. Let $J^{\lambda}_{\pi}(s)$ and $Q^{\lambda}_{\pi}(s,a)$ be the corresponding cost-to-go functions for any $\pi \in \Pi$,
    \begin{align*}
        J^{\lambda}_{\pi}(s) &= \E_{\pi} \left[ \sum_{t=0}^{\infty} \gamma^t g^{\lambda}(s_t,\pi(s_t)) \,\, \Big\vert \,\, s_0=s \right], \\
        Q^{\lambda}_{\pi}(s,a) &= g(s,a) + \gamma \sum_{s' \in \Sc} P(s'|s,a)J^{\lambda}_{\pi}(s').
    \end{align*}
    Similar to \eqref{eq:NPGupdateform}, a quick calculation using the policy gradient theorem reveals that an NPG update for a $\lambda$-regularized MDP solves the following problem,
    \begin{align*}
    \argmin_{\bar{\pi} \in \Pi} \, \langle \nabla \ell^{\lambda}(\pi), \bar{\pi} \rangle \, + \frac{1}{\alpha} \sum_{s \in \Sc} \eta_{\pi}(s) D_{\rm KL}(\bar{\pi}_s || \pi_s)
    \end{align*}
    for any $\alpha \leq 1/\lambda$ with $\langle \nabla \ell^{\lambda}(\pi), \bar{\pi} \rangle = \langle Q^{\lambda}_{\pi} + \lambda \log \pi, \bar{\pi} \rangle_{\eta_{\pi} \times 1}$. For $\alpha = 1/\lambda$, these updates take a particularly simple form of $\pi'(s) = \text{Softmax}\left(\frac{-Q^{\lambda}_{\pi}(s,\cdot)}{\lambda}\right)$.
    This update can alternatively be viewed as a policy iteration update with respect to a regularized Bellman optimality operator, $T^{\lambda}(\cdot)$ defined by:
    \begin{align*}
    (T^{\lambda} J^{\lambda}_{\pi})(s) = \min_{\bar{\pi} \in \Pi} \,\, \langle Q^{\lambda}_{\pi}(s,\cdot), \bar{\pi}(s) \rangle + \lambda \mathcal{H}(\bar{\pi(s)}).
    \end{align*}
    where $\mathcal{H}(\pi(s))=\sum_{i=1}^k \pi(s,i) \log \pi(s,i)$ is the negative entropy. Importantly, $T^{\lambda}(\cdot)$ can also be shown to be a monotone and $\gamma$-contraction in the maximum norm with a unique fixed point, $J^{*, \lambda}$ such that 
    $\|J^{*,\lambda} - J^*\|_{\infty} \leq \frac{\lambda \log k}{(1-\gamma)}$.
    See \citep{geist2019theory} for details. Therefore, similar to the proof of policy iteration in Section \ref{sec: background}, we can obtain a geometric convergence result for NPG with softmax policies and a constant step-size of $\alpha=1/\lambda$,
    \begin{equation}
    \label{eq:regmdpresult}
    \|J_{\pi^t} - J^*\|_{\infty} \leq \gamma^t \|J_{\pi^0} - J^*\|_{\infty} + \frac{2\lambda \log k}{(1-\gamma)^2}.
    \end{equation}
\end{enumerate}

\subsection{Proof of Theorem \ref{thm: tabular rates}}
\begin{proof}
Throughout, we use some standard properties of the Bellman operator as described in Section \ref{sec: background}. We denote $\pi^t_+$ to be the policy iteration update to any policy $\pi^t \in \Pi$ and $\|\cdot\|$ to be the $\ell_{\infty}$-norm.

\paragraph{Part (a): Exact line-search:} Under each algorithm and at each iteration $t$, the policy iteration update $\pi^t_+$ is contained in the class $\Pi^{t+1}$ introduced in Equation \eqref{eq: exact linesearch}.  Therefore, for each algorithm, 
\begin{align}
\label{eq:linesearch_improvement_property}
\ell(\pi^{t+1}) = \min_{\pi \in \Pi^{t+1}}  \ell(\pi)  \leq \ell( \pi^{t}_+) 
\end{align}  
Recall policy improvement property in \eqref{eq:policy_improvement_inequalities_ch3}, which shows $J^* \preceq J_{\pi^t_+} \preceq TJ_{\pi^t} \preceq J_{\pi^t}$. Denote $\rho_{\min} := \min_{s \in \Sc} \rho(s)$. We have,
\begin{align}
	\ell(\pi^t) - \ell(\pi^{t+1}) &\geq \ell(\pi^t) - \ell(\pi^t_{+}) \label{eq:exact_vs_inexact_linesearch} \\
	&= \sum_{s \in \Sc} \rho(s)\left( J_{\pi^t}(s) - J_{\pi^t_+}(s) \right) \nonumber \\
	&\geq \rho_{\min} \left( \sum_{s \in \Sc} J_{\pi^t}(s) - J_{\pi^t_+}(s) \right) \nonumber \\
	&\geq \rho_{\min}\| J_{\pi^t} -  J_{\pi^t_{+}} \| \nonumber \\ 
	&\geq \rho_{\min} \| J_{\pi^t} - T J_{\pi^t} \| \nonumber \\
	&= \rho_{\min} \| \left(J_{\pi^t} - J^*\right) - \left(T J_{\pi^t} - J^*\right) \| \nonumber \\
	&\geq \rho_{\min} \left(\| J_{\pi^t} - J^*\| - \|T J_{\pi^t} - J^* \| \right) \nonumber \\
	&=\rho_{\min} \left( \| J_{\pi^t} - J^*\| - \|T J_{\pi^t} - TJ^*\| \right) \nonumber \\
	&\geq \rho_{\min} \left( 1-\gamma \right)  \| J_{\pi^t} - J^*\| \nonumber \\
	&\geq \rho_{\min} \left( 1-\gamma \right) \left(\ell(\pi^t) - \ell(\pi^*)\right) \nonumber. 
\end{align}
where the second last inequality follows by using the contractivity property of the Bellman operator, that is, $\|TJ_{\pi^t} - TJ^*\| \leq \gamma \|J_{\pi^t} - J^*\|$. Rearranging terms gives, 
\begin{align*} 
\ell(\pi^{t+1}) - \ell(\pi^*) &\leq (1- \rho_{\min}(1-\gamma)  ) \left(\ell(\pi^t)-\ell(\pi^*)\right) \\
&\leq (1- \rho_{\min}(1-\gamma)  )^t \left(\ell(\pi^0)-\ell(\pi^*)\right),
\end{align*}
where the second inequality follows by inductively applying the first one. We immediately have the looser bound $\ell(\pi^{t+1}) - \ell(\pi^*) \leq (1- \rho_{\min}(1-\gamma)  )^t \|  J_{\pi^0} - J^* \|$. The final result follows from observing that
\[
\| J_{\pi^t} - J^*\| \leq \left(\ell(\pi^t) - \ell(\pi^*)\right)/\rho_{\min}.
\]

\paragraph{Part (b): Constant stepsize Frank-Wolfe:}
The proof here follows the analysis of policy iteration reviewed in Section \ref{sec: background}. Recall from Section \ref{sec:algorithms} that a Frank-Wolfe update is equivalent to a soft policy iteration update:
\[
\pi^{t+1}(s) = (1-\alpha)\pi^t(s) + \alpha \pi^t_+(s)
\]
where $\pi^t_+$ is the policy iteration update to $\pi^t$. Thus, starting from a feasible policy $\pi^0 \in \Pi$, we always maintain feasibility for $\alpha \in (0,1]$. By linearity of the cost and transition functions as shown in \eqref{eq: tabular costs and transitions_ch3}, we have that for any state $s$,
\begin{align*}
T_{\pi^{t+1}} J_{\pi^t}(s) 
&= (1-\alpha) J_{\pi^t}(s) + \alpha T_{\pi_+^t} J_{\pi^t}(s) \\
&= (1-\alpha) J_{\pi^t}(s) + \alpha T J_{\pi^t}(s)
\end{align*}
Using $T J_{\pi^t} \preceq J_{\pi^t}$ as in \eqref{eq: bellman update reduces cost_ch3}, we get
\begin{align}
\label{eq:PI_reduces_cost}
T_{\pi^{t+1}} J_{\pi^t} = (1-\alpha) J_{\pi^t} + \alpha T J_{\pi^t} \preceq J_{\pi^t}.
\end{align}
Using monotonicity of $T_{\pi^{t+1}}$, along with the fact that $J_{\pi^{t+1}} =  \lim_{n \to \infty} T^n_{\pi^{t+1}} J_{\pi^t}$ implies,
\[
J_{\pi^t} \succeq T_{\pi^{t+1}} J_{\pi^t} \succeq T^2_{\pi^{t+1}} J_{\pi^t} \succeq \ldots \succeq J_{\pi^{t+1}}
\]
Therefore, from \eqref{eq:PI_reduces_cost}, we get
\[
J_{\pi^{t+1}} \preceq T_{\pi^{t+1}} J_{\pi^t} = (1-\alpha) J_{\pi^t} + \alpha T J_{\pi^t}.
\]
Subtracting $J^*$ from  both sides shows
\[
J_{\pi^{t+1}}- J^* \preceq (1-\alpha)\left(J_{\pi^t} - J^*\right) + \alpha \left(T J_{\pi^t} - J^* \right).
\]
Since the above inequality holds element wise,
\begin{align*}
\norm{J_{\pi^{t+1}} - J^*} &\leq (1-\alpha) \norm{J_{\pi^t} - J^*} +  \alpha \norm{T J_{\pi^t} - J^*} \\
&\leq \left( (1-\alpha) + \gamma \alpha \right) \norm{J_{\pi^t} - J^*},
\end{align*}
where we use that $J^*=TJ^*$ and $\norm{TJ_{\pi^t}-TJ^*}\leq \gamma \norm{J_{\pi^t}-J^*}$ as $T(\cdot)$ is a $\gamma$-contraction. Iterating over the above equation gives us our final result:
\[
\norm{J_{\pi^{t+1}} - J^*} \leq \left(1 - \alpha(1-\gamma) \right)^t \norm{J_{\pi^0} - J^*}.
\]

\paragraph{Part (c): Proof for natural policy gradient with softmax policies and adaptive step-sizes:}
Recall that 
in Section \ref{sec:algorithms}, the natural policy gradient (NPG) update with a step-size sequence $\{\alpha_t\}_{t \geq 0}$ takes the form:
\[
\pi^{t+1}(s,i) = \frac{\pi^{t}(s,i) \cdot e^{-\alpha_t(s) Q^{t}(s,i)}}{\sum_{j=1}^k \pi^{t}(s,j) \cdot e^{-\alpha_t(s) Q^{t}(s,j)}},
\]
where we use the shorthand notation $\pi^{t}(\cdot)$ to denote $\pi_{\theta^{t}}(\cdot)$ and $Q^{t}(s,i)$ to denote $Q_{\pi_{\theta^t}}(s,i)$. For simplicity, we let $c:= \frac{2}{(1-\gamma)}$ which implies, $\alpha_t(s) \geq \frac{c}{\epsilon} \log(\frac{2}{\pi^t(s,i^*_t)})$.

Our proof strategy shows that for any state $s \in \Sc$, an NPG update with step-size $\alpha_t(s)$ decreases the probability of \textit{sub-optimal} actions by a multiplicative factor. Informally, the set of sub-optimal actions per state can be understood to be the set of actions with action gap\footnote{The action gap of any action $i \in \{1,\ldots,k\}$ is the difference between Q-values when compared to the optimal action.} larger than some threshold. Essentially, this shows the NPG update is equivalent to a soft policy iteration update upto a small additive error. We divide the proof into three steps.
	
\paragraph{Step 1: NPG update for \textit{sub-optimal} actions:}
Fix some state $s \in \Sc$. Without loss of generality, we assume the following ordering on the Q-values: $Q^t(s,1) < Q^t(s,2) \ldots < Q^t(s,k)$ which implies that action 1 is optimal in state $s$ under policy $\pi^t$. For error tolerance $\epsilon > 0$, define $O_t^-(s)$ and $O_t^+(s)$ as:
\begin{align*}
O_t^-(s) &:= \left\{ i \,\, |\,\, Q^t(s,i) - Q^t(s,1) \geq \frac{\epsilon}{c} \right\} \\
O_t^+(s) &:= \left\{ i \,\, |\,\, Q^t(s,i) - Q^t(s,1) < \frac{\epsilon}{c} \right\}
\end{align*}
The set $O_t^-(s)$ can be interpreted as the set of \textit{sub-optimal} actions with the \textit{action gap}, $Q^t(s,i) - Q^t(s,1)$, larger than the threshold $\epsilon/c$. Similarly, $O_t^+(s)$ can be interpreted to be the set of \textit{nearly optimal} actions according to policy $\pi^t$. The following lemma (proved in Appendix \ref{appendix}) shows that NPG updates decrease the probability of playing sub-optimal actions by a multiplicative factor.

\begin{restatable}[]{lem}{subopt}
	\label{lemma:suboptimal_actions}
	For any state $s$, $\frac{\pi^{t+1}(s,i)}{\pi^t(s,i)} \leq \frac{1}{2} \,\,\, \forall \, i \in O_t^-(s)$.
\end{restatable}

\paragraph{Step 2: NPG updates as soft policy iteration:}
The policy iteration update, $\pi^t_+(s)=\argmin_{i \in \{1,2,\ldots,k\}} \,\, Q^t(s,i)$, puts entire mass on the best action (according to Q-values of the current policy) and zeros out the probability of playing other actions. On the other hand, Lemma \ref{lemma:suboptimal_actions} shows how an NPG update with appropriate stepsize decays the probabilities of \textit{sub-optimal} actions (in the set $O_t^-(s)$) by a multiplicative factor instead of zeroing them out\footnote{This defintion of sub-optimal actions based on action gap threshold, $\epsilon/c$, is essentially an artifact that we are taking gradient steps with finite step-sizes. As $\alpha_t(s) \to \infty \,\, \forall s \in \Sc$, an NPG update is exactly equal to a policy iteration update.}. This resembles a \textit{soft-policy iteration} update for the set of actions $O_t^-(s)$. We formalize this intuition in the following lemma which characterizes the progress made by an NPG update vis-a-vis a policy iteration update.
\begin{restatable}[Progress quantification]{lem}{progress}
	\label{lemma:progress_quanitification}
	Let $J_{\pi^t}(s)$ denote the cost-to-go function for policy $\pi^t$ from any starting state $s \in \Sc$. Then, 
	\[
	T_{\pi^{t+1}} J_{\pi^t}(s) - J_{\pi^t}(s) \leq  \frac{1}{2} \cdot \left( T J_{\pi^t}(s) - J_{\pi^t}(s) \right) + \frac{\epsilon}{c} 
	\]
\end{restatable}
	
\paragraph{Step 3: Completing the proof:}
Lemma \ref{lemma:progress_quanitification} clearly quantifies the relationship between an NPG update with step-size $\alpha_t$ and a soft policy iteration update with an additive error $\frac{\epsilon}{c}$. With this connection, we give a simple proof of geometric convergence for the natural policy gradient method. First, we claim that $J_{\pi^{t+1}}(s) \leq J_{\pi^t}(s)$. To see this, recall from Section \ref{sec:algorithms} that an NPG update with step-size $\alpha(s)$ can equivalently be written as,
\[
\pi^{t+1}(s) = \argmin_{a \in \Delta^{k-1}} \left[ Q^t(s,a) + \frac{\eta_{\pi^t(s)}}{\alpha(s)} D_{\rm KL}(a||\pi^t(s)) \right]
\]
But staying at the current policy, i.e. taking $a = \pi^t(s)$ is feasible for the optimization problem above. Therefore, 
\[
T_{\pi^{t+1}} J_{\pi^t}(s) = Q^t(s,\pi^{t+1}(s)) \leq Q^t(s,\pi^t(s)) = J_{\pi^t}(s)
\]
Using that $J_{\pi^{t+1}} =  \lim_{n \to \infty} T^n_{\pi^{t+1}} J_{\pi^t}$ along with monotonicity of $T_{\pi^{t+1}}$ implies,
\[
J_{\pi^t} \succeq T_{\pi^{t+1}} J_{\pi^t} \succeq T^2_{\pi^{t+1}} J_{\pi^t} \succeq \ldots \succeq J_{\pi^{t+1}}.
\]
Using this along with Lemma \ref{lemma:progress_quanitification}, we get
\[
J_{\pi^{t+1}} - J_{\pi^t} \preceq T_{\pi^{t+1}}J_{\pi^t} - J_{\pi^t} \preceq \frac{1}{2} \cdot \left( T J_{\pi^t} - J_{\pi^t} \right) + \frac{\epsilon}{c}.
\]
Subtracting $J^*$ from  both sides and rearranging terms gives,
\begin{align*}
J_{\pi^{t+1}}- J^* &\preceq \frac{1}{2} J_{\pi^t} + \frac{1}{2} T J_{\pi^t} - J^* + \frac{\epsilon}{c} \\
&= \frac{1}{2} \left(J_{\pi^t} - J^*\right) +  \frac{1}{2} \left(T J_{\pi^t} - J^* \right) + \frac{\epsilon}{c}.
\end{align*}
As the above inequality holds element wise, we use the contractivity property of $T(\cdot)$ as shown in \eqref{eq:geomteric_contraction_PI_ch3} to get
\begin{align*}
\norm{J_{\pi^{t+1}} - J^*}
\leq \left( \frac{1}{2} + \frac{\gamma}{2} \right) \norm{J_{\pi^t} - J^*} + \frac{\epsilon}{c}.
\end{align*}
Iterating over the above equation and rewriting $\left(\frac{1}{2} + \frac{\gamma}{2}\right) = \left(1-\frac{1}{2}(1-\gamma)\right)$ gives us our desired result.
\end{proof}

\section{Conclusion and Future Work}
In this work, we use illuminating connections with policy iteration as shown in \cite{bhandari2019global} to show how many variants of policy gradient algorithms with large step-sizes and exact gradient evaluations converge geometrically fast for tabular MDPs. An interesting question for future work is whether these results can be extended to function approximation settings where the policy class might be restricted, 
for example in \cite{agarwal2019optimality}. Another interesting question is whether our results hold in settings where unbiased estimates of the value functions are obtained via sampling. Here some exciting progress has been recently made for the undiscounted (average cost setting) in \citep{abbasi2019politex,hao2020provably} for ergodic MDPs, by leveraging connections to approximate policy iteration.            



\newpage

\subsubsection*{Acknowledgements}
We thank anonymous reviewers for their valuable feedback. This work was done in part when JB was participating in the Theory of Reinforcement Learning program at the Simons Institute for the Theory of Computing. JB also thanks Garud Iyengar for his support throughout the PhD program at Columbia University.

\bibliographystyle{plainnat}
\bibliography{bibfile}

\onecolumn
\appendix
{\Large{Appendix: On the Linear Convergence of Policy Gradient Methods for Finite MDPs}}

\vspace{-10pt}
\section{Proof of supporting lemmas}
\label{appendix}
We give proofs of Lemmas \ref{lemma:suboptimal_actions} and \ref{lemma:progress_quanitification}, which were excluded from the main text.

\subopt*
\vspace{-60pt}
\begin{proof}
	The proof follows a simple argument. By definition, for any $i \in O_t^-(s)$:
	\begin{align*}
	\left( Q^t(s,i) - Q^t(s,1) \right) &\geq \frac{\epsilon}{c} \\
	\Rightarrow \, \alpha_t(s) \left( Q^t(s,i) - Q^t(s,1) \right) &\geq \log \frac{2}{\pi^t(s,1)} 
	\end{align*}
	which follows by the definition, $\alpha_t(s) \geq \frac{c}{\epsilon} \log \frac{2}{\pi^t(s,1)}$ which implies $\frac{\epsilon}{c} \geq \frac{1}{\alpha_t(s)} \log \frac{2}{ \pi^t(s,1)}$. Rearranging, we get
	\[
	\log (\pi^{t}(s,1) e^{-\alpha_t(s) Q^{t}(s,1)}) + \log(\frac{1}{2}) \geq -\alpha_t(s) Q^{t}(s,i)
	\]
	Define, $Z_t = \left(\sum_{j=1}^k \pi^{t}(s,j) e^{-\alpha_t(s) Q^{t}(s,j)}\right)$. Then,
	\[
	\log (Z_t) \geq \log \left(\pi^{t}(s,1) e^{-\alpha_t(s) Q^{t}(s,1)} \right)
	\]
	which holds as all the terms in $Z_t$ are positive, i.e. $\pi^{t}(s,j) e^{-\alpha_t(s) Q^{t}(s,j)} > 0 \,\,\, \forall \, j \in \{1,2,\ldots,k\}$, and $\log(\cdot)$ is a monotonic transformation. Rearranging, we get our desired result.
	\begin{align*}
	\log \left( \frac{Z_t}{2}  \right) \geq \log \left(\frac{\pi^t(s,1)}{2} e^{-\alpha_t(s) Q^{t}(s,1)} \right) &\geq -\alpha_t(s) Q^{t}(s,i) \\
	\Rightarrow \,\, \frac{\pi^{t+1}(s,i)}{\pi^t(s,i)} = \frac{1}{Z_t} e^{-\alpha_t(s) Q^{t}(s,i)} &\leq \frac{1}{2}.
	\end{align*}
\end{proof}

\vspace{-50pt}
\progress*
\vspace{-70pt}
\begin{proof}
Fix any state $s \in \Sc$. Without loss of generality, we assume the following ordering on Q-values: $Q^t(s,1) < Q^t(s,2) \ldots < Q^t(s,k)$ which implies that the policy iteration update, $\pi^{+}_t$ puts the entire mass on action 1, which is the best action under the current policy $\pi^t$. That is,
$\pi^t_+(s,1)=1$ and $\pi^t_+(s,i)=0 \,\,\, \forall i \neq 1$. Consider,
	\begin{align}
	T_{\pi^{t+1}} J_{\pi^t}(s) - T J_{\pi^t}(s) &= \langle \pi^{t+1}(s,\cdot) - \pi^t_+(s,\cdot), Q^t(s,\cdot) \rangle \nonumber \\
	&= (\pi^{t+1}(s,1)-1) Q^t(s,1) + \sum_{j=2}^k \pi^{t+1}(s,j) Q^t(s,j) \nonumber \\
	&= - \sum_{j=2}^k \pi^{t+1}(s,j) Q^t(s,1) + \sum_{j=2}^k \pi^{t+1}(s,j) Q^t(s,j) \nonumber \\
	&= \sum_{j=2}^k \pi^{t+1}(s,j) \left( Q^t(s,j) - Q^t(s,1) \right) \nonumber \\
	&= \sum_{j \in \mathcal{O}_{t}^{-}} \pi^{t+1}(s,j) \left( Q^t(s,j) - Q^t(s,1) \right) + \sum_{j \in \mathcal{O}_{t}^{+}} \pi^{t+1}(s,j) \left( Q^t(s,j) - Q^t(s,1) \right) \nonumber \\
	&= \sum_{j \in \mathcal{O}_{t}^{-}} \frac{\pi^{t+1}(s,j)}{\pi^t(s,j)} \pi^t(s,j) \left( Q^t(s,j) - Q^t(s,1) \right) + \sum_{j \in \mathcal{O}_{t}^{+}} \pi^{t+1}(s,j) \underbrace{\left( Q^t(s,j) - Q^t(s,1)\right)}_{< \frac{\epsilon}{c}} \nonumber \\
	&\leq \frac{1}{2} \sum_{j \in \mathcal{O}_{t}^{-}} \pi^t(s,j) \left( Q^t(s,j) - Q^t(s,1) \right) + \frac{\epsilon}{c} \nonumber \\
	&\leq \frac{1}{2} \left( \sum_{j=2}^k \pi^t(s,j) ( Q^t(s,j) - Q^t(s,1) ) \right) + \frac{\epsilon}{c} \nonumber \\
	&= \frac{1}{2} \left( \sum_{j=2}^k \pi^t(s,j) Q^t(s,j) - \sum_{j=2}^k \pi^t(s,j) Q^t(s,1)\right) + \frac{\epsilon}{c} \nonumber \\
	&= \frac{1}{2} \left( \left(\pi^t(s,1)-1\right) Q^t(s,1) + \sum_{j=2}^k \pi^t(s,j) Q^t(s,j) \right) + \frac{\epsilon}{c} \nonumber \\
	&= \frac{1}{2} \, \langle \pi^t(s,\cdot) - \pi^t_+(s,\cdot), Q^t(s,\cdot) \rangle + \frac{\epsilon}{c} \nonumber \\
	&= \frac{1}{2} \, \left( J_{\pi^t}(s) - T J_{\pi^t}(s) \right) + \frac{\epsilon}{c}
	\label{eq:NPG_intermediate}
	\end{align}
	where we used that $\frac{\pi^{t+1}(s,j)}{\pi^t(s,j)} \leq \frac{1}{2} \,\, \forall j \in \mathcal{O}_{t}^{-}(s)$ as shown above in Lemma \ref{lemma:suboptimal_actions} along with the fact that $\left( Q^t(s,j) - Q^t(s,1) \right) \leq \frac{\epsilon}{c} \,\, \forall j \in \mathcal{O}_{t}^{+}(s)$, which follows by definition. Subtracting $J_{\pi^t}(s)$ from both sides in \eqref{eq:NPG_intermediate} and rearranging terms gives our desired result,
	\[
	T_{\pi^{t+1}} J_{\pi^t}(s) - J_{\pi^t}(s) \leq  \frac{1}{2} \cdot \left( T J_{\pi^t}(s) - J_{\pi^t}(s) \right) + \frac{\epsilon}{c}.
	\]
\end{proof}

\section{Details of MDP in Figure \ref{fig:fw_grid}}
\label{appendix_1}
We used the following two state three action MDP, $P \in \mathbb{R}^{|\Sc| |\Ac| \times |\Sc|}, g \in \mathbb{R}^{|\Sc||\Ac|}, \gamma, \rho \in \mathbb{R}^{|\Sc|}$, to generate Figure \ref{fig:fw_grid}. 
\[
P = \begin{bmatrix}
    0.666066 & 0.333934 \\
    0.662211 & 0.337789 \\
    0.441947 & 0.558053 \\
    0.391257 & 0.608743 \\
    0.452186 & 0.547814 \\
    0.035519 & 0.964481
    \end{bmatrix}, \, 
g = \begin{bmatrix}
    0.079718 \\
    0.629733 \\
    0.717644 \\
    0.673362 \\
    0.762623 \\
    0.541251 
    \end{bmatrix}, \, 
\gamma = 0.9, \,
\rho = \begin{bmatrix}
    0.168831 \\
    0.831169
    \end{bmatrix}
\]
Policy $\pi$ for the two states $s_1$ and $s_2$ was taken to be,
\[
\pi(s_1) = \begin{bmatrix}
    0.449416 \\
    0.251788 \\
    0.298796
    \end{bmatrix}, 
\pi(s_2) = \begin{bmatrix}
    0.318626 \\
    0.346284 \\
    0.335090
    \end{bmatrix}.
\]
\end{document}